\documentclass[article]{IEEEtran}

\IEEEoverridecommandlockouts

\usepackage{hyperref}
\usepackage{url}
\usepackage{bbold}
\usepackage{amsfonts}
\usepackage{amsmath,amssymb}
\usepackage{slashbox}

\usepackage{amsthm}
\usepackage{graphicx}
\usepackage{hyperref}
\usepackage{wrapfig}
\usepackage{mathrsfs} 
\usepackage{algorithm,algorithmic}
\usepackage{array}
\usepackage{times}
\usepackage{url}
\usepackage{cite}
\usepackage{upgreek}
\usepackage{bbm}
\usepackage{float}
\usepackage{color}

\usepackage{pgfplots}
\usepackage{color}
\usepackage{flushend}
\usepackage{multirow}
\usepackage{rotating}
\usepackage{tikz,pgfplots}
\usepackage{standalone}

\newcolumntype{C}[1]{>{\centering\let\newline\\\arraybackslash\hspace{0pt}}m{#1}}

\newcommand\numberthis{\addtocounter{equation}{1}\tag{\theequation}}
\newcolumntype{M}[1]{>{\centering\arraybackslash}m{#1}}

\newcommand{\E}{\mathbb{E}}
\newcommand{\g}{\mathbf{G}}
\newcommand{\ba}{\mathbf{B}}
\newcommand{\x}{\bar{\mathbf{x}}}
\newcommand{\sr}{{\tt Succ-Rej}}
\newcommand{\sh}{{\tt Seq-Halv}}
\newcommand{\ns}{{\tt N-Seq-El}}

\providecommand{\abs}[1]{\left\vert#1\right\vert}

\providecommand{\seal}[1]{\left\lceil#1\right\rceil}
\providecommand{\p}[1]{\mathbb{P}\left(#1\right)}
\providecommand{\ex}[1]{\exp\left(#1\right)}
\providecommand{\cl}[1]{\colon\hspace{-0.1cm}#1}

\newtheorem{theorem}{Theorem}
\newtheorem{corollary}[theorem]{Corollary}

\newtheorem{problem}[theorem]{Problem}
\newtheorem{proposition}[theorem]{Proposition}

\newtheorem{fact}{Fact}

\begin{document}

\title{On Sequential Elimination Algorithms for Best-Arm Identification in Multi-Armed Bandits}

%

\author{Shahin Shahrampour, Mohammad Noshad, Vahid Tarokh
\thanks{S. Shahrampour, M. Noshad, and V. Tarokh are with the John A. Paulson School of Engineering and Applied Sciences, Harvard University, Cambridge, MA, 02138 USA. (e-mail: {\tt \{shahin,mnoshad,vahid\}@seas.harvard.edu}).}
}

\maketitle

\begin{abstract}
We consider the best-arm identification problem in multi-armed bandits, which focuses purely on exploration. A player is given a fixed budget to explore a finite set of arms, and the rewards of each arm are drawn independently from a fixed, unknown distribution. The player aims to identify the arm with the largest expected reward. We propose a general framework to unify sequential elimination algorithms, where the arms are dismissed iteratively until a unique arm is left. Our analysis reveals a novel performance measure expressed in terms of the sampling mechanism and number of eliminated arms at each round. Based on this result, we develop an algorithm that divides the budget according to a nonlinear function of remaining arms at each round. We provide theoretical guarantees for the algorithm, characterizing the suitable nonlinearity for different problem environments described by the number of competitive arms. Matching the theoretical results, our experiments show that the nonlinear algorithm outperforms the state-of-the-art. We finally study the side-observation model, where pulling an arm reveals the rewards of its related arms, and we establish improved theoretical guarantees in the pure-exploration setting.
\end{abstract}


\section{Introduction}
Multi-Armed Bandits (MAB) is a sequential decision-making framework addressing the exploration-exploitation dilemma \cite{lai1985asymptotically,auer2002finite,auer2002nonstochastic,bubeck2012regret}. A player explores a finite set of arms sequentially, and pulling each of them results in a {\it reward} to the player. The problem has been studied for different rewards models. In the stochastic MAB, the rewards of each arm are assumed to be i.i.d. samples of an {\it unknown}, fixed distribution. Then, the player's goal is to exploit the arm with largest expected reward as many times as possible to maximize the gain. In the literature, this objective has been translated to minimizing the {\it cumulative} regret, a comparison measure between the actual performance of the player versus a clairvoyant knowing the best arm in advance. Early studies on MAB dates back to a few decades ago, but the problem has surged a lot of renewed interest due to its modern applications. Though generally addressing sequential decision making problems, MAB has been studied in several engineering contexts such as web
search and advertising, wireless cognitive radios, multi-channel communication systems, and Big-Data streaming (see e.g. \cite{mahajan2008multi,liu2010distributed,wang2012optimality,vakili2013deterministic,kalathil2014decentralized,bagheri2015restless,kanoun2016big} and references therein).

Departing from its classical setting (exploration-exploitation), many researchers have studied MAB in the pure-exploration framework. In this case, the player aims to minimize the {\it simple} regret which can be related to finding the best arm with a high probability \cite{audibert2010best}. As a result, the best-arm identification problem has received a considerable attention in the literature of machine learning\cite{even2002pac,mannor2004sample,even2006action,bubeck2009pure,bubeck2011pure,audibert2010best,karnin2013almost,gabillon2012best}. The problem has been viewed from two main standpoints: {\bf (i)} the {\it fixed-confidence} setting, where the objective is to minimize the number of trials to find the best arm with a certain confidence, and {\bf (ii)} the {\it fixed-budget} setting, where the player attempts to maximize the probability of correct identification given a fixed number of arm pulls. Best-arm identification has several applications including channel allocation as originally proposed in \cite{audibert2010best}. Consider the problem of channel allocation for mobile phone communication. Before the outset of communication, a cellphone (player) can explore the set of channels (arms) to find the best one to operate. Each channel feedback is noisy, and the number of trials (budget) is limited. The problem is hence an instance of best-arm identification, and minimizing the cumulative regret is not the right approach to the problem.

While both pure-exploration and exploration-exploitation setups are concerned with finding the best arm, they are quite different problems in nature. In fact, Bubeck et al. \cite{bubeck2011pure} establish that methods designed to minimize the cumulative regret (exploration-exploitation) can perform poorly for the simple-regret minimization (pure-exploration). More specifically, they proved that upper bounds on the cumulative regret yield lower bounds on the simple regret, i.e., the smaller the cumulative regret, the larger the simple regret. Therefore, one must adopt different strategies for optimal best-arm recommendation. 

\subsection{Our Contribution}

In this paper, we address the best-arm identification problem in the fixed-budget setting. We restrict our attention to a class of algorithms that work based on sequential elimination. Recently, it is proved in \cite{carpentier2016tight} that some existing strategies based on the sequential elimination of the arms are {\it optimal}. However, the notion of optimality is defined with respect to the \emph{worst-case} allocation of the reward distributions. The main focus of this paper is not the worst-case scenario. On the contrary, given certain regimes for rewards, our goal is to propose an algorithm outperforming the state-of-the-art in these regimes. We characterize these reward allocations and prove the superiority of our algorithm both theoretically and empirically.

Of particular relevance to the current study is the works of \cite{audibert2010best,karnin2013almost}, where two iterative algorithms are proposed for sequential elimination: Successive Rejects (\sr) \cite{audibert2010best} and Sequential Halving (\sh) \cite{karnin2013almost}. In both algorithms, the player must sample the arms in rounds to discard the arms sequentially, until a single arm is left, one that is perceived as the best arm. However, we recognize two distinctions between the algorithms: {\bf (i)} \sr~eliminates one arm at each round, until it is left with a single arm, whereas \sh~discards roughly half of the remaining arms at each round to identify the best arm. {\bf (ii)} At each round, \sh~samples  the remaining arms uniformly (excluding previous rounds), whereas \sr~samples them uniformly once including the previous rounds. 

Inspired by these works, our first contribution is to propose a general framework to bring sequential elimination algorithms (including \sr~and \sh) under the same umbrella. Our analysis reveals a novel performance bound which relies on the sampling design as well as the number of eliminated arms at each around. Following this general framework, we extend \sr~to an algorithm that divides the budget by a nonlinear function of remaining arms at each round, unlike \sr~that does so in a linear fashion. We prove theoretically that we can gain advantage from the nonlinearity. In particular, we consider several well-studied reward regimes and exhibit the suitable nonlinearity for each environment. Benefiting from the nonlinearity, our algorithm outperforms \sr~and \sh~in these regimes. 
Interestingly, our numerical experiments support our theoretical results, while showing that our algorithm is competitive with {\tt UCB-E} \cite{audibert2010best} which requires prior knowledge of a problem-dependent parameter.

Finally, we consider sequential elimination in the presence of side observations. In this model, a graph encodes the connection between the arms, and pulling one arm reveals the rewards of all neighboring arms \cite{mannor2011bandits}. While the impact of side observations is well-known for exploration-exploitation \cite{caron2012leveraging,buccapatnam2014stochastic}, we consider the model in the pure-exploration setting. Given a partition of arms to a few blocks, we propose an algorithm that eliminates blocks consecutively and selects the best arm from the final block. Naturally, we provide an improved theoretical guarantee comparing to the full bandit setting where there is no side observation.

\subsection{Related Work} 
Pure-exploration in the PAC-learning setup was studied in \cite{even2002pac}, where Successive Elimination for finding an $\epsilon$-optimal arm with probability $1-\delta$ (fixed-confidence setting) was proposed. Seminal works of \cite{mannor2004sample,even2006action} provide matching lower bounds for the problem, which present a sufficient number of arm pulls to reach the confidence $1-\delta$. Many algorithms for pure-exploration are inspired by the classical {\tt UCB1} for exploration-exploitation \cite{auer2002finite}. For instance, Audibert et al. \cite{audibert2010best} propose {\tt UCB-E}, which modifies {\tt UCB1} for pure-exploration. {\tt UCB-E} needs prior knowledge of a problem-dependent parameter, so the authors also propose its adaptive counterpart {\tt AUCB-E} to address the issue. In addition, Jamieson et al. \cite{jamieson2014lil} propose an optimal algorithm for the fixed confidence setting, inspired by the law of the iterated logarithm. We refer the reader to \cite{jamieson2014best} and the references therein for recent advances in the fixed-confidence setting, while remarking that Gabillon et al. \cite{gabillon2012best} present a unifying approach for fixed-budget and fixed-confidence settings. As another interesting direction, various works in the literature introduce information-theoretic measures for best-arm identification. Kaufmann et al. \cite{kaufmann2013information} study the identification of multiple top arms using KL-divergence-based confidence intervals. The authors of \cite{kaufmann2016complexity} investigate both settings to show that the complexity of the fixed-budget setting may be smaller than that of the fixed-confidence setting. Recently, Russo \cite{russo2016simple} develops three Bayesian algorithms to examine asymptotic complexity measure for the fixed-confidence setting. There also exists extensive literature on identification of multiple top arms in MAB (see e.g. \cite{kalyanakrishnan2010efficient,kalyanakrishnan2012pac,kaufmann2013information,bubeck2013multiple,zhou2014optimal,kaufmann2016complexity}). Finally, we remark that simple-regret minimization has been successfully used in the context of Monte-Carlo Tree Search \cite{pepels2014minimizing,liu2015regulation} as well.

\subsection{Organization}
The rest of the paper is organized as follows. Section \ref{Preliminaries} is dedicated to nomenclature, problem formulation, and a summary of our results. In Sections \ref{Nonlinear Sequential Elimination} and  \ref{regimes}, we discuss our main theoretical results and their consequences, while we extend these results to side-observation model in Section \ref{Side Observations}. In Section \ref{simul}, we describe our numerical experiments, and the concluding remarks are provided in Section \ref{Conclusion}. We include the proofs in the Appendix (Section \ref{Appendix}).


\section{Preliminaries}\label{Preliminaries}
{\bf Notation:} For integer $K$, we define $[K]:=\{1,\ldots,K\}$ to represent the set of positive integers smaller than or equal to $K$. We use $\abs{S}$ to denote the cardinality of the set $S$. Throughout, the random variables are denoted in bold letters. 

\begin{table*}[t!]
\caption{The parameters $\alpha$ and $\beta$ for the algorithms discussed in this section, where the misidentification probability for each of them decays in the form of $\beta \ex{- T/\alpha}$. The relevant complexity measures used in this table are defined in \eqref{H2} and \eqref{HP}.}
\begin{center}
\resizebox{14cm}{!}{
\begin{tabular}{| c | c | c | c |} 
 \hline 
  & \sr & \sh &  Our algorithm \\ [.5 ex]
 \hline 
 $\alpha$ & $H_2\overline{\log} K$ & $8H_2 \log_2K$ & $H(p)C_p$ \\ [.5 ex]
 \hline
 $\beta$ & $0.5K(K-1) \ex{K/(H_2\overline{\log} K)}$ & $3 \log_2K$ & $(K-1)\ex{K/H(p)C_p}$ \\ [.5 ex]
 \hline
\end{tabular}}
\end{center}\label{table}
\end{table*}

\subsection{Problem Statement}
Consider the stochastic Multi-armed Bandit (MAB) problem, where a player explores a finite set of $K$ arms. When the player samples an arm, only the corresponding {\it payoff} or {\it reward} of that arm is observed. The reward sequence for each arm $i\in [K]$ corresponds to i.i.d samples of an {\it unknown} distribution whose expected value is $\mu_i$. We assume that the distribution is supported on the unit interval $[0,1]$, and the rewards are generated independently across the arms. Without loss of generality, we further assume that the expected value of the rewards are ordered as 
\begin{align}\label{order}
\mu_1 > \mu_2  \geq \cdots \geq \mu_K,
\end{align}
and therefore, arm $1$ is the unique best arm. We let $\Delta_i:=\mu_1-\mu_i$ denote the {\it gap} between arm $i$ and arm 1, measuring the sub-optimality of arm $i$. We also represent by $\x_{i,n}$ the average reward obtained from pulling arm $i$ for $n$ times. 

In this paper, we address the best-arm identification setup, a pure-exploration problem in which the player aims to find the arm with the largest expected value with a high confidence. There are two well-known scenarios for which the problem has been studied: fixed confidence and fixed budget. In the fixed-confidence setting, the objective is to minimize the number of trials needed to achieve a fixed confidence. However, in this work, we restrict our attention to the fixed-budget, formally described as follows: 
\begin{problem}
Given a total budget of $T$ arm pulls, minimize the probability of misidentifying the best arm.
\end{problem}
It is well-known that classical MAB techniques in the exploration-exploitation setting, such as {\tt UCB1}, are not efficient for the identification of the best arm. In fact, Bubeck et al. have proved in \cite{bubeck2011pure} that upper bounds on the cumulative regret yield lower bounds on the simple regret, i.e., the smaller the cumulative regret, the larger the simple regret. In particular, for all Bernoulli distributions on the rewards, a constant $L>0$, and a function $f(\cdot)$, they have proved if an algorithm satisfies
$$
\text{Expected Cumulative Regret} \leq L f(T), 
$$
after $T$ rounds, we have
\begin{align*}
\text{Misidentification Probability} &\geq \\
\text{Expected Simple Regret} &\geq D_1\exp(-D_2f(T)),
\end{align*}
for two positive constants $D_1$ and $D_2$. Given that for optimal algorithms in the exploration-exploitation setting, we have $f(T)=\mathcal{O}(\log T)$, these algorithms decay polynomially fast for best-arm identification. However, a carefully designed best-arm identification algorithm achieves an exponentially fast decay rate (see e.g. \cite{audibert2010best,karnin2013almost}).

The underlying intuition is that in the exploration-exploitation setting, only playing the best arm matters. For instance, playing the second best arm for a long time can result in a dramatically large cumulative regret. Therefore, the player needs to minimize the exploration time to focus only on the best arm. On the contrary, in the best-arm identification setting, player must recommend the best arm at the end of the game. Hence, exploring the suboptimal arms ``strategically" during the game helps the player to make a better recommendation. In other words, the performance is measured by the final recommendation regardless of the time spent on the suboptimal arms. We focus on sequential elimination algorithms for the best-arm identification in the next section.

\begin{figure*}
\begin{center}
\scalebox{0.96}{\fbox{
\begin{minipage}[t]{8.1cm}
\null 
{\bf \large{General Sequential Elimination Algorithm}}\\

{\bf Input:} budget $T$, sequence $\{z_r,b_r\}_{r=1}^R$.\\
\vspace{-0.24cm}

{\bf Initialize:} $\g_1=[K]$, $n_0=0$.\\
\vspace{-0.24cm}

Let \vspace{-0.32cm}\begin{align*}
C&=\frac{1}{z_R}+\sum_{r=1}^R\frac{b_r}{z_r}\\
n_r&=\seal{\frac{T-K}{Cz_r}} \text{for~~} r \in [R]
\end{align*}

At round $r=1,\ldots,R$:
\begin{itemize}
\item[(1)] Sample each arm in $\g_r$ for $n_r-n_{r-1}$ times. 
\item[(2)] Let $\ba_r$ be the set of $b_r$ arms with smallest average rewards.
\item[(3)] Let $\g_{r+1}=\g_r \setminus \ba_r$, i.e., discard the set worst $b_r$ arms. 
\end{itemize}
{\bf Output:} $\g_{R+1}$.
\end{minipage}}}\scalebox{.96}{\fbox{

\begin{minipage}[t]{8.1cm}
\null
{\bf \large{Nonlinear Sequential Elimination Algorithm}}\\

{\bf Input:} budget $T$, parameter $p>0$.\\
\vspace{-0.24cm}

{\bf Initialize:} $\g_1=[K]$, $n_0=0$.\\
\vspace{-0.24cm}

Let \vspace{-0.32cm}\begin{align*}
C_p&=2^{-p}+\sum_{r=2}^{K}r^{-p}\\
n_r&=\seal{\frac{T-K}{C_p{(K-r+1)}^p}} \text{for~~} r \in [K-1]
\end{align*}

At round $r=1,\ldots,K-1$:
\begin{itemize}
\item[(1)] Sample each arm in $\g_r$ for $n_r-n_{r-1}$ times. 
\item[(2)] Let $\ba_r$ be the set containing the arm with smallest average reward.
\item[(3)] Let $\g_{r+1}=\g_r \setminus \ba_r$, i.e., discard the worst single arm. 
\end{itemize}
{\bf Output:} $\g_{K}$.
\end{minipage}}}
\caption{The algorithm on the left represents a general recipe for sequential elimination, whereas the one on the right is a special case of the left hand side, which extends \sr~to nonlinear budget allocation.}
\label{ALGO}
\end{center}
\end{figure*}

\subsection{Previous Performance Guarantees and Our Result}
In this work, we examine sequential-elimination type algorithms in the fixed budget setting. We propose a general algorithm that unifies sequential elimination methods, including celebrated \sr~\cite{audibert2010best} and \sh~\cite{karnin2013almost}. We then use a special case of this general framework to develop an algorithm called Nonlinear Sequential Elimination. We show that this algorithm is more efficient than \sr~and \sh~in several problem scenarios. 

Any sequential elimination algorithm samples the arms based on some strategy. It then discards a few arms at each round and stops when it is only left by one arm. In order to integrate sequential elimination algorithms proceeding in $R$ rounds, we use the following key observation: let any such algorithm play each (remaining) arm for $n_r$ times (in total) by the end of round $r \in [R]$. If the algorithm needs to discard $b_r$ arms at round $r$, it must satisfy the following budget constraint 
\begin{align}\label{budget1}
b_1n_1+b_2n_2+\cdots+(b_R+1)n_R \leq T,
\end{align}
since the $b_r$ arms eliminated at round $r$ have been played $n_r$ times, and the surviving arm has been played $n_R$ times. Alternatively, letting $g_r:=\sum_{i=r}^R b_i+1$ denote the number of remaining arms at the start of round $r$, one can pose the budget constraint as
\begin{align}\label{budget1}
g_1n_1+g_2(n_2-n_1)+\cdots+g_R(n_R-n_{R-1}) \leq T.
\end{align}
Our first contribution is to derive a generic performance measure for such algorithm (Theorem \ref{mainthm}), relating the algorithm efficiency to $\{b_r\}_{r=1}^R$ as well as the sampling scheme determining $\{n_r\}_{r=1}^R$. Note that \sr~satisfies $b_r=1$ for $R=K-1$ rounds, whereas \sh~is characterized via $g_{r+1}=\seal{g_r/2}$ with $g_1=K$ for $R=\seal{\log_2K}$. While it is shown in \cite{mannor2004sample} that in many settings for MAB, the quantity $H_1$ in the following plays a key role, 
\begin{align}\label{H2}
H_1:=\sum_{i=2}^K \frac{1}{\Delta_i^2}  \ \ \ \ \  \ \ \  \text{and}  \ \ \ \ \ \  \ \ H_2:=\max_{i \neq 1} \frac{i}{\Delta_i^2},
\end{align}
the performance of both \sr~and \sh~is described via the complexity measure $H_2$,  
which is equal to $H_1$ up to logarithmic factors in $K$ \cite{audibert2010best}. In particular, for each algorithm the bound on the probability of misidentification can be written in the form of $\beta \ex{- T/\alpha}$, where $\alpha$ and $\beta$ are given in Table \ref{table} in which $\overline{\log}~K=0.5+\sum_{i=2}^Ki^{-1}$. 

 In \sr, at round $r$, the $K-r+1$ remaining arms are played proportional to the whole budget divided by $K-r+1$ which is a linear function of $r$. Motivated by the fact that this linear function is not necessarily the best sampling rule, as our second contribution, we specialize our general framework to an algorithm which can be cast as a nonlinear extension of \sr. This algorithm is called Nonlinear Sequential Elimination (\ns), where the term ``nonlinear'' refers to the fact that at round $r$, the algorithm divides the budget by the nonlinear function $(K-r+1)^p$ for a positive real $p>0$ (an input value). We prove (Proposition \ref{mainprop}) that the performance of our algorithm depends on the following quantities
\begin{align}\label{HP}
H(p):=\max_{i \neq 1} \frac{i^p}{\Delta_i^2} \  \ \ \  \ \  \text{and}   \ \ \ \  \ \ C_p:=2^{-p}+\sum_{r=2}^{K}r^{-p},
\end{align}
as described in Table \ref{table}. We do not use $p$ as a subscript in the definition of $H(p)$ to avoid confusion over the fact that $H(1)=H_2$ due to definition \eqref{H2}. Indeed, \ns~with $p=1$ recovers \sr, but we show that in many regimes for arm gaps, $p\neq 1$ provides better theoretical results (Corollary \ref{ratecorollary}). We also illustrate this improvement in the numerical experiments in Section \ref{simul}, where we observe that $p\neq 1$ can outperform \sr~and \sh~in many settings considered in \cite{audibert2010best,karnin2013almost}. Since the value of $p$ is received as an input, we remark that our algorithm needs tuning to perform well; however, the tuning is more qualitative rather than quantitative, i.e., the algorithm maintains a reasonable performance as long as $p$ is in a certain interval, and therefore, the value of $p$ needs not be specific. We will discuss this in the next section in more details.


\section{Nonlinear Sequential Elimination}\label{Nonlinear Sequential Elimination}
In this section, we propose our generic sequential elimination method and analyze its performance in the fixed budget setting. Consider a sequential elimination algorithm given budget  $T$ of arm pulls. The algorithm maintains an active set initialized by the $K$ arms, and it proceeds for $R$ rounds to discard the arms sequentially until it is left by a single arm. Let us use $\seal{\cdot}$ to denote the ceiling function. Then, for a constant $C$ and a decreasing, positive sequence $\{z_r\}_{r=1}^R$, set $n_r= \seal{(T-K)/(Cz_r)}$ at round $r\in [R]$, let the algorithm sample the remaining arms for $n_r-n_{r-1}$ times, and calculate the empirical average of rewards for each arm. If the algorithm dismisses $b_r$ arms with lowest average rewards, and we impose the constraint $\sum_{r=1}^Rb_r=K-1$ on the sequence $\{b_r\}_{r=1}^R$, the algorithm outputs a single arm, the one that it hopes to be the best arm. This general point of view is summarized in Fig. \ref{ALGO} on the left side.

The choice of $C$ in the algorithm (see Fig. \ref{ALGO}) must warrant that the budget constraint \eqref{budget1} holds. When we substitute $n_r$ into \eqref{budget1}, we get
\begin{align*}
&n_R+\sum_{r=1}^Rb_rn_r = \seal{\frac{T-K}{Cz_R}} +\sum_{r=1}^R b_r \seal{\frac{T-K}{Cz_r}} \\
&~~~~~~~~~~~~~~~~ \leq \frac{T-K}{Cz_R} +1+ \sum_{r=1}^R b_r +\sum_{r=1}^R b_r \frac{T-K}{Cz_r} \\
&~~~~~~~~~~~~~~~~ = K + \frac{T-K}{C} \left(\frac{1}{z_R} +\sum_{r=1}^R \frac{b_r}{z_r}\right)=T,
\end{align*}
where in the last line we used the condition $\sum_{r=1}^R b_r=K-1$. The following theorem provides an upper bound on the error probability of the algorithm.

\begin{theorem}\label{mainthm}
Consider the General Sequential Elimination algorithm outlined in Fig. \ref{ALGO}. For any $r\in [R]$, let $b_r$ denote the number of arms that the algorithm eliminates at round $r$. Let also $g_r:=\abs{\g_r}=\sum_{i=r}^R b_i+1$ be the number of remaining arms at the start of round $r$. Given a fixed budget $T$ of arm pulls and the input sequence $\{z_r\}_{r=1}^R$, setting $C$ and $\{n_r\}_{r=1}^R$ as described in the algorithm, the misidentification probability satisfies the bound, 
\begin{align*} 
&\p{\g_{R+1}\neq\{1\}}\leq \\
& ~~~~~~~~~ R \max_{r\in [R]}\{b_r\}\ex{-\frac{T-K}{C}\min_{r\in [R]}\left\{\frac{2\Delta_{g_{r+1}+1}^2}{z_r}\right\}}.
\end{align*}
\end{theorem}
It is already quite well-known that the sub-optimality $\Delta_i$ of each arm $i$ plays a key role in the identification quality; however, an important subsequent of Theorem \ref{mainthm} is that the performance of any sequential elimination algorithm also relies on the choice of $z_r$ which governs the constant $C$. In \cite{audibert2010best}, \sr~employs $z_r=K-r+1$, i.e., at each round the remaining arms are played equally often in total. This results in $C$ being of order $\log K$.

We now use the abstract form of the generic algorithm to specialize it to Nonlinear Sequential Elimination delineated in Fig. \ref{ALGO}. The algorithm works with $b_r=1$ and $z_r={(K-r+1)}^p$ for a given $p>0$, and it is called ``nonlinear'' since $p$ is not necessarily equal to one. The choice of $p=1$ reduces the algorithm to \sr. In Section \ref{regimes}, we prove that in many regimes for arm gaps, $p\neq 1$ provides better theoretical results, and we further exhibit the efficiency in the numerical experiments in Section \ref{simul}. The following proposition encapsulates the theoretical guarantee of the algorithm. 
\begin{proposition}\label{mainprop}
Let the Nonlinear Sequential Elimination algorithm in Fig. \ref{ALGO} run for a given $p>0$, and let $C_p$ and $H(p)$ be as defined in \eqref{HP}. Then, the misidentification probability satisfies the bound, 
$$
\p{\g_{K}\neq\{1\}} \leq (K-1)\ex{-2\frac{T-K}{C_pH(p)}}.
$$ 
\end{proposition}
The performance of the algorithm does depend on the parameter $p$, but the choice is more qualitative rather than quantitative. For instance, if the sub-optimal arms are almost the same, i.e. $\Delta_i \approx \Delta$ for $i\in [K]$, noting the definition of $C_p$ and $H(p)$ in \eqref{HP}, we observe that $0<p<1$ performs better than $p>1$. In general, larger values for $p$ increase $H(p)$ and decrease $C(p)$. Therefore, there is a trade-off in choosing $p$. We elaborate on this issue in Sections \ref{regimes} and \ref{simul}, where we observe that a wide range of values for $p$ can be used for tuning, and the trade-off can be addressed using either $0<p<1$ or $1<p\leq 2$.

We remark that in the proof of Theorem \ref{mainthm}, the constant behind the exponent can be improved if one can avoid the union bounds. To do so, one needs to assume some structure in the sequence $\{g_r\}_{r=1}^R$. For instance, the authors in \cite{karnin2013almost} leverage the fact that $g_{r+1}=\seal{g_r/2}$ to avoid union bounds in \sh. The idea can be extended to when the ratio of $g_{r+1}/g_r$ is a constant independent of $r$. 

Finally, while the result of Proposition \ref{mainprop} provides a general performance bound with respect to reward regimes, in the next section, we provide two important corollaries of the proposition to study several regimes for rewards, where we can simplify the bound and compare our result with other algorithms.


\section{Performance in Several Sub-optimality Regimes}\label{regimes}
Inspired by numerical experiments carried out in the previous works \cite{audibert2010best,karnin2013almost}, we consider a few instances for sub-optimality of arms in this section, and we demonstrate how \ns~fares in these cases. We would like to distinguish three general regimes that encompass interesting settings for sub-optimality and determine the values of $p$ for which we expect \ns~to achieve faster identification rates: 
\begin{itemize}
\item[{\bf 1}] {\bf Arithmetic progression of gaps:} In this case, $\Delta_i=(i-1)\Delta_0$ for $i>1$, where $\Delta_0$ is a constant. 
\item[{\bf 2}] {\bf A large group of competitive, suboptimal arms:} We recognize this case as when there exists a constant\footnote{The choice of $\varepsilon$ must be constant with respect to $i$ and $K$.} $\varepsilon\geq 0$ such that $\Delta_i/\Delta_2\leq 1+\varepsilon$ for arms $i \in S$, where $\abs{S}$ grows linearly as a function of $K$, and for $i \notin S$, $\Delta_i/\Delta_2\geq i$. 
\item[{\bf 3}] {\bf A small group of competitive, suboptimal arms:} This case occurs when there exists a constant $\varepsilon \geq 0$ such that $\Delta_i/\Delta_2\leq 1+\varepsilon$ for $i \in S$, where $\abs{S}$ is of constant order with respect to $K$, and for $i \notin S$, $\Delta_i/\Delta_2\geq i$.
\end{itemize}
We now state the following corollary of Proposition \ref{mainprop}, which proves to be useful for our numerical evaluations. Note that the orders are expressed with respect to $K$.
\begin{corollary}\label{ratecorollary}
Consider the Nonlinear Sequential Elimination algorithm described in Fig. \ref{ALGO}. Let constants $p$ and $q$ be chosen such that $1<p\leq2$ and $0<q<1$. Then, for the three settings given above, the bound on the misidentification probability presented in Proposition \ref{mainprop} satisfies
\begin{table}[h!]
\begin{center}
\resizebox{\columnwidth}{!}{
\begin{tabular}{ | c | c | c |} 
 \hline 
   {\bf Regime 1} &    {\bf Regime 2} &     {\bf Regime 3} \\ [.5 ex]
 \hline 
 $H(p)C_p=\mathcal{O}(1)$ & $H(q)C_q=\mathcal{O}(K)$ & $H(p)C_p=\mathcal{O}(1)$ \\ [.5 ex]
 \hline
\end{tabular}}
\end{center}
\end{table}
\end{corollary} 
We can now compare our algorithm with \sr~and \sh~ using the result of Corollary \ref{ratecorollary}. Returning to Table \ref{table} and calculating $H_2$ for Regimes $1$ to $3$, we can observe the following table, 
\begin{table}[h!]
\caption{The misidentification probability for all three algorithms decays in the form of $\beta \ex{- T/\alpha}$. The table represents the parameter $\alpha$ for each algorithm in Regimes 1 to 3. For Regime 2, we set $0<p<1$, and for Regimes 1 and 3, we use $1< p \leq 2$.}
\begin{center}
\resizebox{\columnwidth}{!}{
\begin{tabular}{| c | c | c | c |} 
 \hline 
  & \sr & \sh &  Our algorithm \\ [.5 ex]
 \hline 
 {\bf Regime 1} & $\mathcal{O}(\log K)$ & $\mathcal{O}(\log K)$ & $\mathcal{O}(1)$ \\ [.5 ex]
 \hline
{\bf Regime 2} & $\mathcal{O}(K\log K)$ & $\mathcal{O}(K\log K)$ & $\mathcal{O}(K)$ \\ [.5 ex]
 \hline
{\bf Regime 3} & $\mathcal{O}(\log K)$ & $\mathcal{O}(\log K)$ & $\mathcal{O}(1)$ \\ [.5 ex]
 \hline
\end{tabular}}\label{table2}
\end{center}
\end{table}
which shows that with a right choice of $p$ for \ns, we can save a $\mathcal{O}(\log K)$ factor in the exponential rate comparing to other methods. Though we do not have a prior information on gaps to categorize them into one of the Regimes $1$ to $3$ (and then choose $p$), the choice of $p$ is more qualitative rather than quantitative. Roughly speaking: if the sub-optimal arms are almost the same $0<p<1$ performs better than $p>1$, and if there are a few real competitive arms, $p>1$ outperforms $0<p<1$. Therefore, the result of Corollary \ref{ratecorollary} is of practical interest, and we will show using numerical experiments (Section \ref{simul}) that a wide range of values for $p$ can potentially result in efficient algorithms. 

One should observe that the number of competitive arms is a key factor in tuning $p$. We now provide another corollary of Proposition \ref{mainprop}, which presents the suitable range of parameter $p$ given the growth rate of  competitive arms as follows.
\begin{corollary}\label{ratecorollary2}
Let the number of competitive arms be an arbitrary function $f_K$ of total arms, i.e., there exists a constant $\varepsilon \geq 0$ such that $\Delta_i/\Delta_2\leq 1+\varepsilon$ for $i \in S$, where $\abs{S}=f_K$, and for $i \notin S$, $\Delta_i/\Delta_2\geq i$. Then, there exists a suitable choice of $p$ for which \ns~outperforms \sr~and \sh~in the sense that the misidentification probability decays with a faster rate, i.e., we have $\mathcal{O}(H(p)C_p) \leq \mathcal{O}(H_2 \log K) $. For different conditions on the growth of $f_K$, the suitable choice of parameter $p$ is presented in Table \ref{table3}.
\begin{table}[h!]
\caption{Given that the misidentification probability decays in the form of $\beta \ex{- T/\alpha}$, the table represents the choice of $p$ for which the parameter $\alpha$ in \ns~is smaller than those of \sr~and~\sh.}
\begin{center}
\resizebox{\columnwidth}{!}{
\begin{tabular}{ | >{\centering\arraybackslash}m{1.12in} | >{\centering\arraybackslash}m{1.9in} | @{}m{0pt}@{}} 
 \hline 
  {\bf Condition on $f_K$}  &   {\bf Suitable Range of Parameter $p$} & \\ [2ex]
 \hline 
 $1 \leq f_K\leq \log K$ & $1<p\leq 2$ & \\ [3.5 ex]
 \hline
 $\log K < f_K<  \frac{K}{\log K}$ & $1-\frac{\log\log K}{\log \left(\frac{K}{f_K}\right)}<p<1+\frac{\log\log K}{\log f_K}$& \\ [3.5 ex]
 \hline
 $ \frac{K}{\log K} \leq f_K \leq K-1$ & $0<p< 1$& \\ [3.5 ex]
 \hline
\end{tabular}}
\end{center}\label{table3}
\end{table}
\end{corollary}
The corollary above indicates that a side information in the form of the number of competitive arms (in the order) can help us tune the input parameter $p$. The corollary exhibits a smooth interpolation between when the competitive arms are small versus when they are large. Perhaps, the most interesting regime is the middle row, where the choice of $p$ is given as a function of the number of arms $K$. Consider the following example where we calculate the choice of $p$ for $f_K=K^\gamma$, where $\gamma \in\{0.3,0.5,0.7\}$. 
\begin{table}[h!]
\caption{The table shows the suitable interval of $p$ for $K\in\{40,120,5*10^2,5*10^4,5*10^6,5*10^8\}$ given the growth rate of competitive arms as derived in Corollary \ref{ratecorollary2}.}
\begin{center}
\resizebox{\columnwidth}{!}{
\begin{tabular}{| >{\centering\arraybackslash}m{.6in} | >{\centering\arraybackslash}m{.6in} | >{\centering\arraybackslash}m{.6in} | >{\centering\arraybackslash}m{.6in} | @{}m{0pt}@{}} 
 \hline 
 \backslashbox{$K$}{$f_K$} &  $K^{0.3}$ &  $K^{0.5}$ &   $K^{0.7}$ \\ [.5 ex]
  \hline
  $40$ & $(0.5,2)$ & $(0.3,1.7)$ & $(0,1.5)$ &\\ [1.5 ex]
  \hline
$120$ & $(0.53,2)$ & $(0.35,1.65)$ & $(0,1.47)$ &\\ [1.5 ex]
 \hline
  $5*10^2$ & $(0.58,1.97)$ & $(0.42,1.58)$ & $(0.03,1.42)$ &\\ [1.5 ex]
  \hline
$5*10^4$ & $(0.69,1.73)$ & $(0.56,1.44)$ & $(0.27,1.31)$ &\\ [1.5 ex]
 \hline
 $5*10^6$ & $(0.75,1.59)$ & $(0.65,1.35)$ & $(0.41,1.25)$ &\\ [1.5 ex]
 \hline
 $5*10^8$ & $(0.79,1.5)$ & $(0.7,1.3)$ & $(0.5,1.21)$ &\\ [1.5 ex]
 \hline
 \end{tabular}}
\end{center}
\end{table}

As we can see in the table above, since the condition on $p$ depends in logarithmic orders on $K$ and $f_K$, we have flexibility to tune $p$ even for very large $K$. However, we only use $K\in \{40,120\}$ for the numerical experiments in Section \ref{simul}, since the time-complexity of Monte Carlo simulations is prohibitive on large number of arms.

\section{Side Observations}\label{Side Observations}
\begin{figure*}[t!]
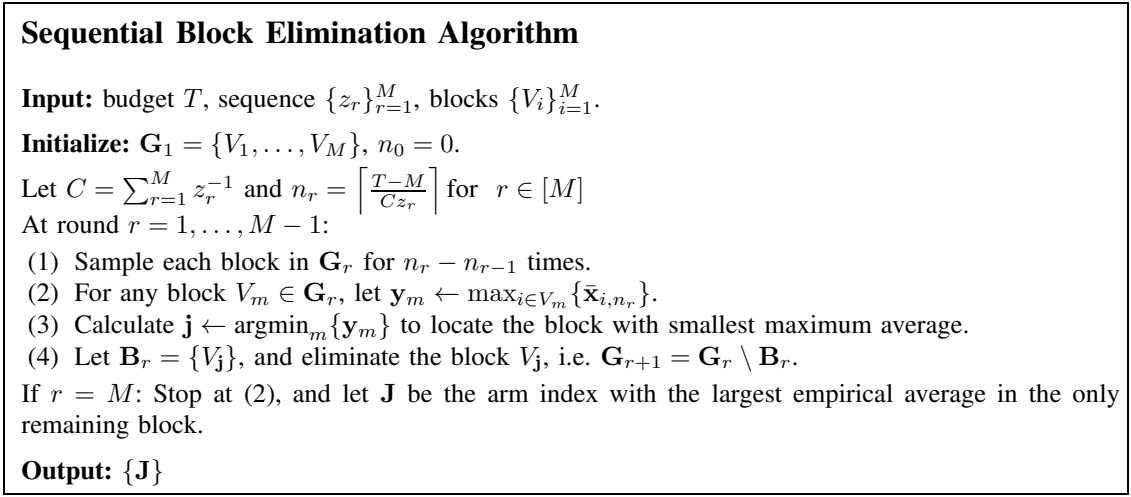

\begin{center}
\scalebox{1}{\fbox{
\begin{minipage}[t]{14.6cm}
\null 
{\bf \large{Sequential Block Elimination Algorithm}}\\

{\bf Input:} budget $T$, sequence $\{z_r\}_{r=1}^M$, blocks $\{V_i\}_{i=1}^M$.\\
\vspace{-0.24cm}

{\bf Initialize:} $\g_1=\{V_1,\ldots,V_M\}$, $n_0=0$.\\
\vspace{-0.24cm}

Let  $C=\sum_{r=1}^Mz_r^{-1}$ and $n_r=\seal{\frac{T-M}{Cz_r}} \text{for~~} r \in [M]$

At round $r=1,\ldots,M-1$:
\begin{itemize}
\item[(1)] Sample each block in $\g_r$ for $n_r-n_{r-1}$ times. 
\item[(2)] For any block $V_m \in  \g_r$, let $\mathbf{y}_m \leftarrow \max_{i\in V_m}\{\x_{i,n_r}\}$.
\item[(3)] Calculate $\mathbf{j} \leftarrow \text{argmin}_m\{\mathbf{y}_m\}$ to locate the block with smallest maximum average.
\item[(4)] Let $\ba_r=\{V_\mathbf{j}\}$, and eliminate the block $V_\mathbf{j}$, i.e. $\g_{r+1}=\g_r \setminus \ba_r$.
\end{itemize}
If $r=M$: Stop at (2), and let $\mathbf{J}$ be the arm index with the largest empirical average in the only remaining block.\\
\vspace{-0.24cm}

{\bf Output:} $\{\mathbf{J}\}$
\end{minipage}}}
\caption{Sequential Block Elimination discards the blocks one by one and selects the best arm in the last block.}
\label{ALGO2}
\end{center}
\end{figure*}

 In the previous sections, we considered a scenario in which pulling an arm yields only the reward of the chosen arm. However, there exist applications where pulling an arm can additionally result in some side observations. For a motivative example, consider the problem of web advertising, where an ad placer offers an ad to a user and receives a reward only if the user clicks on the ad. In this example, if the user clicks on a vacation ad, the ad placer receives the side information that the user could have also clicked on ads for rental cars. The value of side observations in the stochastic MAB was studied in \cite{caron2012leveraging} for exploration-exploitation setting. The side-observation model is described via an undirected graph that captures the relationship between the arms. Once an arm is pulled, the player observes the reward of the arm as well as its neighboring arms. In exploration-exploitation settings, the analysis of MAB with side observations relies on the cliques of the graph \cite{mannor2011bandits,caron2012leveraging}. In this section, we would like to consider the impact in the pure-exploration setting. 

In the pure-exploration, we minimize the simple regret rather than the cumulative regret, and therefore, the player's best bets are the most connected arms resulting in more observations. Now consider a partition of the set $[K]$ into $M$ blocks $\{V_i\}_{i=1}^M$ such that each $V_i$ contains a star graph, i.e., there exists an arm in $V_i$ connected to all other arms. Given such partition, the player can follow a simple rule to leverage side observations and eliminate blocks one by one. The idea is to sample the central arm in each block, which reveals the rewards for all the arms in that block.  At round $r$, sample the remaining blocks (the central arms of each block) for $n_r-n_{r-1}$ times, and find the arm with largest average reward in each block. With the best arms of all blocks at hand, remove the block whose best arm is the worst comparing to other blocks. Continue the sampling until only one block is left, and output the arm with largest average reward in the final block as the best arm. The algorithm, called  Sequential Block Elimination, is summarized in Fig. \ref{ALGO2}, and its performance is characterized in the following theorem.

\begin{theorem}\label{sidethm}
Let the Sequential Block Elimination algorithm in Fig. \ref{ALGO2} run given the input sequence $\{z_r\}_{r=1}^M$ and blocks $\{V_i\}_{i=1}^M$. Define $V:=\max_{i\in [M]}\{\abs{V_i}\}$ to be the maximum cardinality among the blocks. Then, the misidentification probability satisfies the bound 
$$
\p{\{\mathbf{J}\} \neq \{1\}} \leq VM \ex{-\frac{T-M}{C}2\min_{r\in [M]}\left\{\frac{\Delta_{M+1-r}^2}{z_r}\right\}}.
$$
In particular, the choice of $z_r={(M+1-r)}^p$ for a given $p>0$, yields 
$$
\p{\{\mathbf{J}\} \neq \{1\}} \leq VM \ex{-2\frac{T-M}{CH_{M,p}}},
$$
where $H(M,p):=\max_{i\in [M]\setminus \{1\}}\left\{\frac{i^p}{\Delta_i^2}\right\}$.
\end{theorem}
The theorem indicates that once we can partition the arms into $M$ blocks, the complexity measure $H(p)$ need not be maximized over $K$ arms. Instead, it is maximized over the top $M$ arms. The improvement must be more visible in the settings that arms are roughly as competitive as each other, since we can slow down the linear growth of $K$ by $M$. 

As a final note, $M=K$ recovers the fully bandit (no side observation) setting with $V=1$, and we observe that in such case, we recover the result of Proposition \ref{mainprop} from Theorem \ref{sidethm} in the rate of exponential decay. However, the constant behind the exponential would change from $K-1$ in Proposition \ref{mainprop} to $K$ in Theorem \ref{sidethm}. This is the artifact of an extra term contributing to the upper bound, which can be removed for the case $M=K$.


\section{Numerical Experiments}\label{simul}

\begin{figure*}[t!]
\begin{center}
        \includegraphics[scale=1.06]{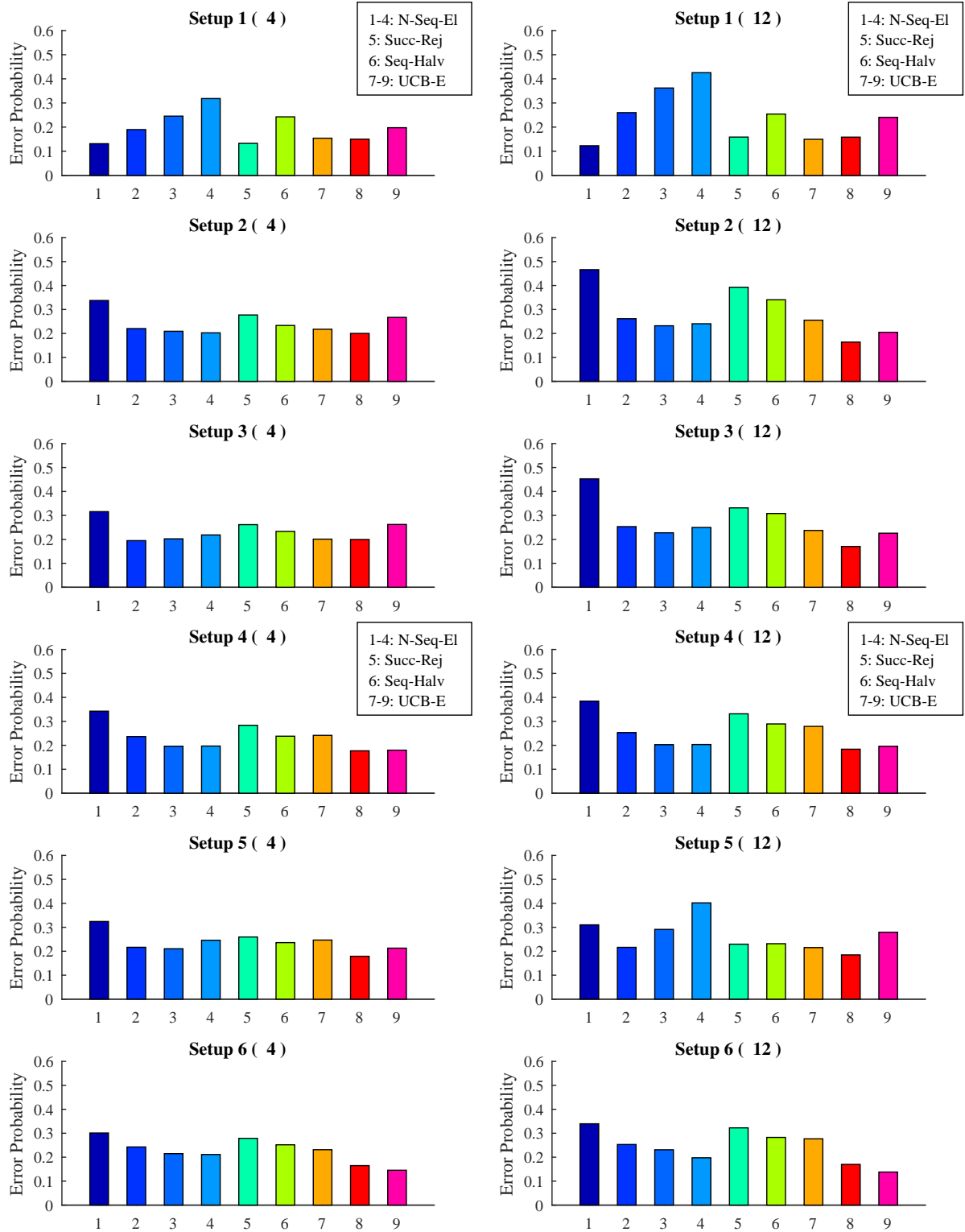}
\end{center}
\vspace{-1.79cm}
	\caption{The figure shows the misidentification probability for \ns, \sr, \sh, and {\tt UCB-E} algorithms in six different setups. The six plots on the left hand side relate to the case $K=40$, and the six plots on the right hand side are associated with $K=120$. The height of each bar depicts the misidentification probability, and each index (or color) represents one algorithm tuned with a specific parameter in case the algorithm is parameter dependent.}
	\label{plot}
\end{figure*}

In this section, we empirically evaluate our algorithm on the settings perviously studied in \cite{audibert2010best,karnin2013almost}. In these experiments, we compare \ns~with \sr, \sh, and {\tt UCB-E}. Though we include {\tt UCB-E} proposed in \cite{audibert2010best} as a benchmark, we remark that the algorithm requires a prior knowledge of a parameter that depends on $H_1$ defined in \eqref{H2}. The adaptive version of {\tt UCB-E} was also developed in \cite{audibert2010best}. The algorithm (called {\tt AUCB-E}) does not need prior knowledge of $H_1$, and it calculates the parameter online. The experiments in \cite{karnin2013almost} suggest that for $T\approx H_1$, \sr, \sh, and {\tt UCB-E} outperform {\tt AUCB-E}, and it is not surprising that the prior knowledge of $H_1$ must give {\tt UCB-E} an advantage over {\tt AUCB-E}.

We consider Bernoulli distribution on the rewards, assuming that the expected value of Bernoulli distribution for the best arm is $\mu_1=0.7$. In what follows, we use the notation $x\cl{y}$ to denote integers in $[x,y]$. We examine the following setups for two values of arm numbers $K \in \{40,120\}$:
\begin{itemize}
\item[{\bf 1}]{\bf One group of suboptimal arms:} $\mu_{2\cl{K}}=0.6$.
\item[{\bf 2}]{\bf Two groups of suboptimal arms:} $\mu_{2:m}=0.7-\frac{2}{K}$, $\mu_{m+1:K}=0.4$, and $m=\seal{\log \frac{K}{2}+1}$. 
\item[{\bf 3}]{\bf Three groups of suboptimal arms:} $\mu_{2:m}=0.7-\frac{2}{K}$, $\mu_{m+1:2m}=0.7-\frac{4}{K}$, $\mu_{2m+1:K}=0.4$, and $m=\seal{\log \frac{K}{2}+1}$. 
\item[{\bf 4}]{\bf Arithmetic progression:} $\Delta_i=\frac{0.6(i-1)}{K-1}$ for $i=2\cl{K}$.
\item[{\bf 5}]{\bf Geometric progression:} $\Delta_i=0.01(1+\frac{4}{K})^{i-2}$ for $i=2\cl{K}$.
\item[{\bf 6}]{\bf One real competitive arm:} $\mu_{2}=0.7-\frac{1}{2K}$ and $\mu_{3:K}=0.2$.
\end{itemize}
We run $4000$ experiments for each setup with specific $K$, and the misidentification probability is averaged out over the experiment runs. The budget $T$ considered in each setup is equal to $\seal{H_1}$ in the corresponding setup following \cite{audibert2010best,karnin2013almost}. Fig. \ref{plot} illustrates the performance of each algorithm in different setups. The height of each bar depicts the misidentification probability, and the index guideline is as follows: {\bf (i)} indices 1-4: \ns~with parameter $p=0.75,1.35,1.7,2$. {\bf (ii)} index 5: \sr. {\bf (iii)} index 6: \sh. {\bf (iv)} indices 7-9: {\tt UCB-E} with parameter $a=cT/H_1$, for $c=1,2,4$. The legends are the same for all the plots, and we remove some of them to avoid clutter.

The results are consistent with Corollary \ref{ratecorollary}, and the following comments are in order:
\begin{itemize}
\item Setup 1 perfectly relates to Regime 2 in Corollary \ref{ratecorollary} ($\varepsilon=0$). While any choice of $0<p<1$ gives an $\mathcal{O}(K)$ rate, the rate deteriorates to $\mathcal{O}(K^p)$ by choosing $1<p\leq 2$. Therefore,  only the the choice of $p=0.75$ is appropriate for \ns. This choice joined with \sr~(which amounts to our algorithm with $p=1$) outperform others even {\tt UCB-E}.
\item Setup 4 defines Regime 1 in Corollary \ref{ratecorollary}. Accordingly, choice of $1<p\leq 2$ outperforms \sr~and \sh, and $p=1.7,2$ prove to be competitive to {\tt UCB-E}. 
\item In Setups 2-3, the number of competitive arms grows with $\mathcal{O}(\log K)$, but they can be considered close to Regime 3 in Corollary \ref{ratecorollary} as the growth is sub-linear. Also, Setup 6 is the ideal match for Regime 3 in Corollary \ref{ratecorollary}. Therefore, any choice of $1<p\leq 2$ should be suitable in these cases. We observe that for $p=1.35,1.7,2$, \ns~outperforms \sr~and \sh, while being quite competitive to {\tt UCB-E} in Setups 2-3 for $K=40$.
\item In Setup 5, we were expecting to observe good performance for $1<p\leq 2$. While we do see that for $K=40$, \ns~outperforms \sr~and \sh~with $p=1.35,1.7$, the performance is not quite satisfactory for $K=120$ (only $p=1.35$ performs well). A potential reason is that we want to keep the rewards bounded in $[0,1]$, thereby choosing the geometric rate for $\Delta_i$ so slow that it is dominated by ${i}^p$ rate in $\eqref{HP}$. To support our argument, we present a complementary evaluation of this case for a small $K$, where we can use faster geometric growth. 
\item As a final remark, note that in consistent with Table \ref{table2}, for relevant cases we observe an improvement of performance when $K$ increases. For instance, in Setup 1 with $p=0.75$, the ratio of misidentification probability of \ns~to \sh~increases from $1.84~(K=40)$ to $2.06~(K=120)$, or in Setup 4 with $p=2$, the ratio of misidentification probability of \ns~to \sr~increases from $1.43~(K=40)$ to $1.63~(K=120)$.
\end{itemize}

\subsection{Complementary Numerical Experiment for Geometric Sub-optimality Setup}
\begin{figure}[t]
\begin{center}
        \includegraphics[scale=0.62]{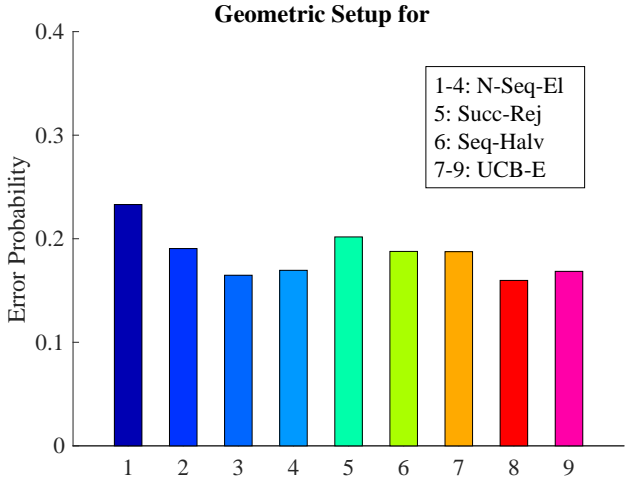}
\end{center}
	\caption{The figure shows the misidentification probability for \ns, \sr, \sh, and {\tt UCB-E} algorithms for an experiment with $K=7$. The height of each bar depicts the misidentification probability, and each index (or color) represents one algorithm tuned with a specific parameter in case the algorithm is parameter dependent.}
	\label{plot2}
\end{figure}

As we remarked in the very last comment of numerical experiments, choosing small number of arms, we can space the expected value of arm rewards such that in \eqref{HP} the sub-optimality term $\Delta_i$ dominates $i^p$ for $1<p\leq 2$. We propose a simple experiment for $K=7$ arms, where $\mu_1=0.7$ and $\Delta_i=(0.6)^{8-i}$ in Fig. \ref{plot2}. As expected, \ns~tuned with $p>1$ is competitive, and it achieves its best performance with $p=1.7$, winning the competition against others, while almost equalizing with {\tt UCB-E} for $c=2$.

\section{Conclusion}\label{Conclusion}
We considered best-arm identification in the stochastic multi-armed bandits, where a player is given a certain budget to explore a finite number of arms. The player's objective is to detect the arm with largest expected reward. We contribute to the literature of best-arm identification by {\bf (i)} unifying sequential elimination algorithms under a general framework, which introduces a novel performance metric for this class of algorithms,
{\bf (ii)} developing a nonlinear sequential elimination algorithm with provable theoretical and practical guarantees, {\bf (iii)} and establishing a theoretical result on the value of side observations in the pure-exploration setting.

Having established that we gain advantage from nonlinear budget allocation, an important future direction is to propose a method that starts with a specific nonlinear rate and fine-tunes the rate according to the problem environment. The main challenge is that the quantity $C$ should perhaps be time-varying, and its value needs to be cautiously controlled by the algorithm, so that the algorithm does not overspend the budget. 

\section*{Acknowledgments}
We gratefully acknowledge the support of DARPA under grant numbers N66001-15-C-4028 and W911NF-14-1-0508. 
We thank Prof. Masahiro Yukawa (Keio University) for his helpful comments. We also thank Dr. Ahmad Beirami and Dr. Hamed Farhadi for many helpful discussions.


\section{Appendix}\label{Appendix}
\begin{fact}(Hoeffding's inequality) \label{HOFF}
Let $W_1,\ldots,W_n$ be independent random variables with support on the unit interval with probability
one. If $S_n=\sum_{i=1}^nW_i$, then for all $a>0$, it holds that
$$
\p{S_n-\E[S_n]\geq a} \leq \ex{\frac{-2a^2}{n}}. 
$$
\end{fact}

\section*{Proof of Theorem \ref{mainthm}}
Recall that $\g_r$ denotes the set of arms not eliminated by the start of round $r$ with $g_r=\abs{\g_r}$ being its cardinality. Also, $\ba_r$ represents the set of arms that we decide to discard after playing round $r$ with $b_r=\abs{\ba_r}$ denoting its cardinality. It evidently holds that $\g_{r+1}=\g_r \setminus \ba_r$ for $r\in [R]$. Therefore,
\begin{align}\label{eq:1}
\p{\g_{R+1}\neq\{1\}}=\p{1\in \cup_{r=1}^R \ba_r}=\sum_{r=1}^R \p{1\in \ba_r},
\end{align}
since the sets of removed arms at each round are disjoint, i.e. $\ba_i \cap \ba_j =\emptyset$ for $i\neq j$. We can then write
\begin{align}\label{eq:2}
 \p{1\in \ba_r}= \sum_{G_r}\p{1\in \ba_r ~ \vert ~ \g_r=G_r}\p{\g_r=G_r}.
\end{align}
Now for any particular $G_r$, consider the worst $b_r$ arms, i.e., the bottom $b_r$ arms when arms are ordered in terms of (true) expected value.    If the best arm (arm $1$) is set to be eliminated at the end of round $r$, its empirical average must be less than at least one of these $b_r$ arms. In the case that $G_r=\{1,2,\ldots,g_r\}$, the bottom $b_r$ arms would be $\{g_r-b_r+1,\ldots,g_r\}$. Therefore, recalling that $\x_{i,n}$ denotes the average reward of pulling arm $i$ for $n$ times, and using Hoeffding's inequality (Fact \ref{HOFF}), we get
\begin{align*}
&\p{1\in \ba_r ~ \vert ~ \g_r=\{1,2,\ldots,g_r\}} \\
&~~~~~~~~~~~~~\leq \sum_{i=g_r-b_r+1}^{g_r} \p{\x_{1,n_r} \leq \x_{i,n_r}} \\  
&~~~~~~~~~~~~~\leq \vphantom{\sum_{i=g_r-b_r+1}^{g_r} \p{\x_{1,n_r} \leq \x_{i,n_r}}}b_r \ex{-2n_r\Delta_{g_r-b_r+1}^2}\\
&~~~~~~~~~~~~~=b_r \ex{-2n_r\Delta_{g_{r+1}+1}^2}, \label{eq:3} \numberthis
\end{align*}
where the last step is due to the fact that $g_{r+1}=g_r-b_r$. In any other case for $G_r$ the best of the worst $b_r$ arms cannot be better than arm $g_r-b_r+1$. As a result, combining \eqref{eq:2} and \eqref{eq:3}, we obtain
\begin{align*}
 \p{1\in \ba_r} &\leq \sum_{G_r} b_r \ex{-2n_r\Delta_{g_{r+1}+1}^2} \p{\g_r=G_r}\\
 &=b_r \ex{-2n_r\Delta_{g_{r+1}+1}^2}.
\end{align*}
Then, in view of \eqref{eq:1} we derive
\begin{align*}
&\p{\g_{R+1}\neq\{1\}}\leq \sum_{r=1}^R b_r \ex{-2n_r\Delta_{g_{r+1}+1}^2} \\
&~~~~~~~~~~\leq R \max_{r\in [R]}\{b_r\}\max_{r\in [R]}\left\{\ex{-2n_r\Delta_{g_{r+1}+1}^2}\right\}.
\end{align*}
Noting the fact that $n_r=\seal{\frac{T-K}{Cz_r}} \geq \frac{T-K}{Cz_r}$, we can use above to conclude that
\begin{align*}
&\p{\g_{R+1}\neq\{1\}} \leq \\
&~~~~~~~~~~~~~R \max_{r\in [R]}\{b_r\}\max_{r\in [R]}\left\{\ex{-\frac{T-K}{Cz_r}2\Delta_{g_{r+1}+1}^2}\right\},
\end{align*}
which completes the proof. \qed

\section*{Proof of Proposition \ref{mainprop}}
We point out that the algorithm is a special case of General Sequential Elimination where $R=K-1$, $b_r=1$, $g_r=K-r+1$, and $z_r={(K-r+1)}^p$. The proof then follows immediately from the result of Theorem \ref{mainthm}. \qed

\section*{Proof of Corollary \ref{ratecorollary}}
The proof follows by substituting each case in \eqref{HP}. We need to understand the order of $C_p$ and $H(p)$ for different regimes of $p$. Let us start by
$$
C_p=2^{-p}+\sum_{r=2}^{K}r^{-p},
$$
and noting that for any $p>1$, $C_p$ is a convergent sum when $K\rightarrow \infty$. Therefore, for the regime $p>1$, the sum is a constant, i.e., $C_p=\mathcal{O}(1)$. On the other hand, consider $q \in (0,1)$, and note that the sum is divergent, and for large $K$ we have $C_q = \mathcal{O}(K^{1-q})$. Now, let us analyze
$$
H(p)=\max_{i \neq 1} \frac{i^p}{\Delta_i^2}.
$$
For Regime 1, since $\Delta_i=(i-1)\Delta_0$, for $p\in (1,2)$, we have 
\begin{align*}
H(p)&=\max_{i \neq 1} \left(1+\frac{1}{i-1}\right)^p\frac{(i-1)^{p-2}}{\Delta_0^2}\\
&\leq \max_{i \neq 1} \left(1+\frac{1}{i-1}\right)^p\frac{1}{\Delta_0^2}=\frac{1.5^p}{\Delta_0^2},
\end{align*}
 which is of constant order with respect to $K$. Therefore, the product $C_pH(p)=\mathcal{O}(1)$. 
 
 For Regime 2, we have
 $$
H(q)=\max_{i \neq 1} \frac{i^q}{\Delta_i^2} \leq \max_{i \neq 1} \frac{i^q}{\Delta_2^2} \leq \frac{K^q}{\Delta_2^2},
$$
and the maximum order can be achieved as the number of arms close to the second best arm grows linearly in $K$.
Combining with $C_q = \mathcal{O}(K^{1-q})$, the product $C_qH(q) =\mathcal{O}(K)$. 

For Regime 3, if $i\in S$, we have 
$$
\max_{i \in S} \frac{i^p}{\Delta_i^2}=\mathcal{O}(1),
$$
since the cardinality of $S$ is of constant order with respect to $K$. On the other hand, since $1<p\leq 2$, we have
$$
\max_{i \notin S} \frac{i^p}{\Delta_i^2} \leq \max_{i \notin S} \frac{i^p}{i^2\Delta_2^2}=\max_{i \notin S} \frac{i^{p-2}}{\Delta_2^2}=\mathcal{O}(1).
$$
Therefore, $H(p)$ is of constant order, and combining with $C_p=\mathcal{O}(1)$, the product $C_pH(p)=\mathcal{O}(1)$.
 \qed

\section*{Proof of Corollary \ref{ratecorollary2}}
According to calculations in the previous proof, we have that $C_p = \mathcal{O}(K^{1-p})$ for $0<p<1$, and $C_p = \mathcal{O}(1)$ for $1<p\leq 2$. In order to find out the order of $H(p)$ and $H(2)$, we should note that  
$$
\mathcal{O}(H(p))=\mathcal{O}\left(\max_{i \in S} \frac{i^p}{\Delta_i^2}\right) = \mathcal{O}\left(\max_{i \in S} \frac{i^p}{\Delta_2^2}\right)= f_K^p,
$$
and we simply have $\mathcal{O}(H_2)=\mathcal{O}(H(1))=f_K$. Now returning to Table \ref{table}, we must compare $\mathcal{O}(H(p)C_p)=f_K^p\mathcal{O}(C_p)$ and $\mathcal{O}(H_2 \log K)=f_K\log K$ for all three the rows given in Table \ref{table3}. 

{\bf Row 1:} In the case that $1 \leq f_K\leq \log K$, for any choice of $1 < p \leq 2$, since $C_p=\mathcal{O}(1)$, we always have 
$$
f_K^p \leq f_K \log K \Leftrightarrow  f_K^{p-1} \leq \log K.
$$

{\bf Row 3:} In the case that $ \frac{K}{\log K} \leq f_K \leq K-1$, for any choice of $0 < p < 1$, since $C_p = \mathcal{O}(K^{1-p})$, we have
\begin{align*}
f_K^p\mathcal{O}(C_p)= f_K^pK^{1-p}=f_K \left(\frac{K}{f_K}\right)^{1-p} \leq f_K \log K.
\end{align*}

{\bf Row 2:} In this case, we have that $\log K < f_K<  \frac{K}{\log K}$. To prove the claim in the Table \ref{table3}, we have to break down the analysis into two cases:

{\bf Row 2 -- Case 1:} First, consider the case $1-\frac{\log\log K}{\log \left(\frac{K}{f_K}\right)}<p<1$. Since $C_p = \mathcal{O}(K^{1-p})$, we have that
\begin{align*}
f_K^p\mathcal{O}(C_p)&= f_K^pK^{1-p}=f_K \left(\frac{K}{f_K}\right)^{1-p} \\
&\leq f_K \left(\frac{K}{f_K}\right)^{\frac{\log\log K}{\log \left(\frac{K}{f_K}\right)}}=f_K \log K.
\end{align*}

{\bf Row 2 -- Case 2:} Second, consider the case $1<p<1+\frac{\log\log K}{\log f_K}$. Since $C_p=\mathcal{O}(1)$, we have that
\begin{align*}
f_K^p\mathcal{O}(C_p)&= f_K^p\leq f_K ^{1+\frac{\log\log K}{\log f_K}}= f_K f_K ^{\frac{\log\log K}{\log f_K}}=f_K \log K. 
\end{align*}
Therefore, for all of the conditions on $f_K$, we showed proper choice of $p$ which guarantees $f_K^p\mathcal{O}(C_p) \leq f_K \log K$.  \qed

\section*{Proof of Theorem \ref{sidethm}}
Recall that the elements of $\g_r$ are the set of $M+1-r$ blocks not eliminated by the start of round $r \in [M]$, and $\ba_r$ only contains the block we decide to discard after playing round $r$. Also, recall that arm $1$ is the best arm, located in the block $V_1$, and $V:=\max_{i\in [M]}\{\abs{V_i}\}$ denotes the maximum cardinality of the blocks. 

 If the algorithm does not output the best arm, the arm is either eliminated with block $V_1$ in one the rounds in $[M-1]$, or not selected at the final round $M$. Therefore,
\begin{align*}
\p{\mathbf{J} \neq 1}&\leq \p{\text{argmax}_{i \in V_1}\{\x_{i,n_M}\} \neq 1 ~ \vert ~ \g_M=\{V_1\} }\\
&+\sum_{r=1}^{M-1} \p{\ba_r=\{V_1\}}.  \label{eq:11} \numberthis
\end{align*}
The first term can be bounded simply by Hoeffding's inequality 
\begin{align*}
&\p{\text{argmax}_{i \in V_1}\{\x_{i,n_M}\} \neq 1 ~ \vert ~ \g_M=\{V_1\} } \leq \\
&~~~~~~~~~~~\vphantom{\sum_{i=1}^t}\abs{V_1} \ex{-2n_M\Delta_2^2}=\abs{V_1} \ex{-2n_M\Delta_1^2}, \label{eq:12} \numberthis
\end{align*}
using the convention $\Delta_1:=\Delta_2$. On the other hand, for any $r\in [M-1]$
\begin{align*}
 &\p{\ba_r=\{V_1\}}= \\
 &~~~~~~~~~\vphantom{\sum_{i=1}^t}\sum_{G_r}\p{\ba_r=\{V_1\} ~ \vert ~ \g_r=G_r}\p{\g_r=G_r}. \numberthis \label{eq:13}
\end{align*}
Note that, without loss of generality, we sort the blocks based on their best arm as 
\begin{align*}
\mu_1=\max_{i\in V_1}{\mu_i} > \max_{i\in V_2}{\mu_i} \geq \cdots \geq \max_{i\in V_M}{\mu_i},
\end{align*} 
so the best possible arm in $V_m$ cannot be better than arm $m$ in view of above and \eqref{order}. 
We remove block $V_1$ after execution of round $r$ only if it is the worst among all other candidates.
Therefore, consider the particular case that $G_r=\{V_1,V_2,\ldots,V_{M+1-r}\}$ contains the best possible $M+1-r$ blocks that one can keep until the start of round $r$. In such case, 
\begin{align*}
&\p{\ba_r=\{V_1\} ~ \vert ~ \g_r=\{V_1,V_2,\ldots,V_{M+1-r}\}}\\
 &~~~~~~~~~~~~~~~~~~~~~~~~~~~~~~~~~~~~\leq V\ex{-2n_r\Delta_{M+1-r}^2}.
\end{align*}
In any other case for $G_r$ the best possible arm in the worst block cannot be better than arm $M+1-r$. Therefore, combining above with \eqref{eq:13}, we obtain 
\begin{align*}
 \p{\ba_r=\{V_1\}}&\leq V \sum_{G_r}\ex{-2n_r\Delta_{M+1-r}^2}\p{\g_r=G_r}\\
 &=V\ex{-2n_r\Delta_{M+1-r}^2}.
\end{align*}
Incorporating above and \eqref{eq:12} into \eqref{eq:11}, we derive
\begin{align*}
\p{\mathbf{J} \neq 1}& \leq \vphantom{V\sum_{r=1}^M\ex{-n_r\Delta^2_{M+1-r}}}\abs{V_1} \ex{-2n_M\Delta_1^2}\\
&+\sum_{r=1}^{M-1}V\ex{-2n_r\Delta^2_{M+1-r}}\\
& \leq V\sum_{r=1}^M\ex{-2n_r\Delta^2_{M+1-r}}\\
& \leq VM \ex{-\frac{T-M}{C}2\min_{r\in [M]}\left\{\frac{\Delta_{M+1-r}^2}{z_r}\right\}},
\end{align*}
noticing the choice of $n_r$ in the algorithm (Fig. \ref{ALGO2}). Having finished the first part of the proof, when we set $z_r={(M+1-r)}^p$ for $r\in [M-1]$ and $z_M=2^{p}$, we have
$$
\p{\mathbf{J} \neq 1} \leq VM \ex{-2\frac{T-M}{CH_{M,p}}},
$$
where $H(M,p):=\max_{i\in [M]\setminus \{1\}}\left\{\frac{i^p}{\Delta_i^2}\right\}$.\qed

\bibliographystyle{IEEEtran}
\bibliography{IEEEabrv,references}

\begin{thebibliography}{10}
\providecommand{\url}[1]{#1}
\csname url@samestyle\endcsname
\providecommand{\newblock}{\relax}
\providecommand{\bibinfo}[2]{#2}
\providecommand{\BIBentrySTDinterwordspacing}{\spaceskip=0pt\relax}
\providecommand{\BIBentryALTinterwordstretchfactor}{4}
\providecommand{\BIBentryALTinterwordspacing}{\spaceskip=\fontdimen2\font plus
\BIBentryALTinterwordstretchfactor\fontdimen3\font minus
  \fontdimen4\font\relax}
\providecommand{\BIBforeignlanguage}[2]{{%
\expandafter\ifx\csname l@#1\endcsname\relax
\typeout{** WARNING: IEEEtran.bst: No hyphenation pattern has been}%
\typeout{** loaded for the language `#1'. Using the pattern for}%
\typeout{** the default language instead.}%
\else
\language=\csname l@#1\endcsname
\fi
#2}}
\providecommand{\BIBdecl}{\relax}
\BIBdecl

\bibitem{lai1985asymptotically}
T.~L. Lai and H.~Robbins, ``Asymptotically efficient adaptive allocation
  rules,'' \emph{Advances in applied mathematics}, vol.~6, no.~1, pp. 4--22,
  1985.

\bibitem{auer2002finite}
P.~Auer, N.~Cesa-Bianchi, and P.~Fischer, ``Finite-time analysis of the
  multiarmed bandit problem,'' \emph{Machine learning}, vol.~47, no. 2-3, pp.
  235--256, 2002.

\bibitem{auer2002nonstochastic}
P.~Auer, N.~Cesa-Bianchi, Y.~Freund, and R.~E. Schapire, ``The nonstochastic
  multiarmed bandit problem,'' \emph{SIAM Journal on Computing}, vol.~32,
  no.~1, pp. 48--77, 2002.

\bibitem{bubeck2012regret}
S.~Bubeck and N.~Cesa-Bianchi, ``Regret analysis of stochastic and
  nonstochastic multi-armed bandit problems,'' \emph{Machine Learning}, vol.~5,
  no.~1, pp. 1--122, 2012.

\bibitem{mahajan2008multi}
A.~Mahajan and D.~Teneketzis, ``Multi-armed bandit problems,'' in
  \emph{Foundations and Applications of Sensor Management}.\hskip 1em plus
  0.5em minus 0.4em\relax Springer, 2008, pp. 121--151.

\bibitem{liu2010distributed}
K.~Liu and Q.~Zhao, ``Distributed learning in multi-armed bandit with multiple
  players,'' \emph{IEEE Transactions on Signal Processing}, vol.~58, no.~11,
  pp. 5667--5681, 2010.

\bibitem{wang2012optimality}
K.~Wang and L.~Chen, ``On optimality of myopic policy for restless multi-armed
  bandit problem: An axiomatic approach,'' \emph{IEEE Transactions on Signal
  Processing}, vol.~60, no.~1, pp. 300--309, 2012.

\bibitem{vakili2013deterministic}
S.~Vakili, K.~Liu, and Q.~Zhao, ``Deterministic sequencing of exploration and
  exploitation for multi-armed bandit problems,'' \emph{IEEE Journal of
  Selected Topics in Signal Processing}, vol.~7, no.~5, pp. 759--767, 2013.

\bibitem{kalathil2014decentralized}
D.~Kalathil, N.~Nayyar, and R.~Jain, ``Decentralized learning for multiplayer
  multiarmed bandits,'' \emph{IEEE Transactions on Information Theory},
  vol.~60, no.~4, pp. 2331--2345, 2014.

\bibitem{bagheri2015restless}
S.~Bagheri and A.~Scaglione, ``The restless multi-armed bandit formulation of
  the cognitive compressive sensing problem,'' \emph{IEEE Transactions on
  Signal Processing}, vol.~63, no.~5, pp. 1183--1198, 2015.

\bibitem{kanoun2016big}
K.~Kanoun, C.~Tekin, D.~Atienza, and M.~Van Der~Schaar, ``Big-data streaming
  applications scheduling based on staged multi-armed bandits,'' \emph{IEEE
  Transactions on Computers}, 2016.

\bibitem{audibert2010best}
J.-Y. Audibert and S.~Bubeck, ``Best arm identification in multi-armed
  bandits,'' in \emph{COLT-23th Conference on Learning Theory-2010}, 2010, pp.
  13--p.

\bibitem{even2002pac}
E.~Even-Dar, S.~Mannor, and Y.~Mansour, ``{PAC} bounds for multi-armed bandit
  and markov decision processes,'' in \emph{Computational Learning
  Theory}.\hskip 1em plus 0.5em minus 0.4em\relax Springer, 2002, pp. 255--270.

\bibitem{mannor2004sample}
S.~Mannor and J.~N. Tsitsiklis, ``The sample complexity of exploration in the
  multi-armed bandit problem,'' \emph{The Journal of Machine Learning
  Research}, vol.~5, pp. 623--648, 2004.

\bibitem{even2006action}
E.~Even-Dar, S.~Mannor, and Y.~Mansour, ``Action elimination and stopping
  conditions for the multi-armed bandit and reinforcement learning problems,''
  \emph{The Journal of Machine Learning Research}, vol.~7, pp. 1079--1105,
  2006.

\bibitem{bubeck2009pure}
S.~Bubeck, R.~Munos, and G.~Stoltz, ``Pure exploration in multi-armed bandits
  problems,'' in \emph{Algorithmic Learning Theory}.\hskip 1em plus 0.5em minus
  0.4em\relax Springer, 2009, pp. 23--37.

\bibitem{bubeck2011pure}
------, ``Pure exploration in finitely-armed and continuous-armed bandits,''
  \emph{Theoretical Computer Science}, vol. 412, no.~19, pp. 1832--1852, 2011.

\bibitem{karnin2013almost}
Z.~Karnin, T.~Koren, and O.~Somekh, ``Almost optimal exploration in multi-armed
  bandits,'' in \emph{Proceedings of the 30th International Conference on
  Machine Learning (ICML-13)}, 2013, pp. 1238--1246.

\bibitem{gabillon2012best}
V.~Gabillon, M.~Ghavamzadeh, and A.~Lazaric, ``Best arm identification: A
  unified approach to fixed budget and fixed confidence,'' in \emph{Advances in
  Neural Information Processing Systems}, 2012, pp. 3212--3220.

\bibitem{carpentier2016tight}
A.~Carpentier and A.~Locatelli, ``Tight (lower) bounds for the fixed budget
  best arm identification bandit problem,'' in \emph{29th Annual Conference on
  Learning Theory}, 2016, pp. 590--604.

\bibitem{mannor2011bandits}
S.~Mannor and O.~Shamir, ``From bandits to experts: On the value of
  side-observations,'' in \emph{Advances in Neural Information Processing
  Systems}, 2011, pp. 684--692.

\bibitem{caron2012leveraging}
S.~Caron, B.~Kveton, M.~Lelarge, and S.~Bhagat, ``Leveraging side observations
  in stochastic bandits,'' \emph{Uncertainty in Artificial Intelligence (UAI)},
  2012.

\bibitem{buccapatnam2014stochastic}
S.~Buccapatnam, A.~Eryilmaz, and N.~B. Shroff, ``Stochastic bandits with side
  observations on networks,'' in \emph{The 2014 ACM international conference on
  Measurement and modeling of computer systems}, 2014, pp. 289--300.

\bibitem{jamieson2014lil}
K.~Jamieson, M.~Malloy, R.~Nowak, and S.~Bubeck, ``lil'ucb: An optimal
  exploration algorithm for multi-armed bandits,'' in \emph{Proceedings of The
  27th Conference on Learning Theory}, 2014, pp. 423--439.

\bibitem{jamieson2014best}
K.~Jamieson and R.~Nowak, ``Best-arm identification algorithms for multi-armed
  bandits in the fixed confidence setting,'' in \emph{Information Sciences and
  Systems (CISS), 2014 48th Annual Conference on}.\hskip 1em plus 0.5em minus
  0.4em\relax IEEE, 2014, pp. 1--6.

\bibitem{kaufmann2013information}
E.~Kaufmann and S.~Kalyanakrishnan, ``Information complexity in bandit subset
  selection,'' in \emph{Conference on Learning Theory}, 2013, pp. 228--251.

\bibitem{kaufmann2016complexity}
E.~Kaufmann, O.~Capp{\'e}, and A.~Garivier, ``On the complexity of best-arm
  identification in multi-armed bandit models,'' \emph{Journal of Machine
  Learning Research}, vol.~17, no.~1, pp. 1--42, 2016.

\bibitem{russo2016simple}
D.~Russo, ``Simple bayesian algorithms for best arm identification,'' in
  \emph{29th Annual Conference on Learning Theory}, 2016.

\bibitem{kalyanakrishnan2010efficient}
S.~Kalyanakrishnan and P.~Stone, ``Efficient selection of multiple bandit arms:
  Theory and practice,'' in \emph{Proceedings of the 27th International
  Conference on Machine Learning (ICML-10)}, 2010, pp. 511--518.

\bibitem{kalyanakrishnan2012pac}
S.~Kalyanakrishnan, A.~Tewari, P.~Auer, and P.~Stone, ``{PAC} subset selection
  in stochastic multi-armed bandits,'' in \emph{Proceedings of the 29th
  International Conference on Machine Learning (ICML-12)}, 2012, pp. 655--662.

\bibitem{bubeck2013multiple}
S.~Bubeck, T.~Wang, and N.~Viswanathan, ``Multiple identifications in
  multi-armed bandits,'' in \emph{Proceedings of The 30th International
  Conference on Machine Learning}, 2013, pp. 258--265.

\bibitem{zhou2014optimal}
Y.~Zhou, X.~Chen, and J.~Li, ``Optimal {PAC} multiple arm identification with
  applications to crowdsourcing,'' in \emph{Proceedings of the 31st
  International Conference on Machine Learning (ICML-14)}, 2014, pp. 217--225.

\bibitem{pepels2014minimizing}
T.~Pepels, T.~Cazenave, M.~H. Winands, and M.~Lanctot, ``Minimizing simple and
  cumulative regret in monte-carlo tree search,'' in \emph{Computer
  Games}.\hskip 1em plus 0.5em minus 0.4em\relax Springer, 2014, pp. 1--15.

\bibitem{liu2015regulation}
Y.-C. Liu and Y.~Tsuruoka, ``Regulation of exploration for simple regret
  minimization in monte-carlo tree search,'' in \emph{IEEE Conference on
  Computational Intelligence and Games (CIG)}, 2015, pp. 35--42.

\end{thebibliography}

\end{document}